\documentclass[10pt,twocolumn,letterpaper]{article}

\usepackage{iccv}              % To produce the CAMERA-READY version
\usepackage{multirow}
\usepackage{tabularx}
\usepackage{makecell}
\usepackage{colortbl}
\definecolor{graycolor}{rgb}{0.92,0.92,0.92}
\definecolor{iccvblue}{rgb}{0.21,0.49,0.74}
\usepackage[pagebackref,breaklinks,colorlinks,allcolors=iccvblue]{hyperref}

\def\paperID{7042} % *** Enter the Paper ID here
\def\confName{ICCV}
\def\confYear{2025}

\title{RobuSTereo: Robust Zero-Shot Stereo Matching under Adverse Weather}

\author{Yuran Wang\textsuperscript{1}\thanks{These authors contributed equally.} \quad Yingping Liang\textsuperscript{1}$^{*}$ \quad Yutao Hu\textsuperscript{2} \quad Ying Fu\textsuperscript{1}\thanks{Corresponding author: fuying@bit.edu.cn} \\
\textsuperscript{1}Beijing Institute of Technology \qquad \textsuperscript{2}School of Computer Science and Engineering, Southeast University \\
{\tt\small\{wangyuran,liangyingping,fuying\}@bit.edu.cn \quad huyutao@seu.edu.cn }
}
\begin{document}
\maketitle
% \twocolumn[{%
% \renewcommand\twocolumn[1][]{#1}%
% \maketitle
% \begin{center}
%     \centering
%     \includegraphics[width=\textwidth]{img/ts-all.png}
%     % \subfloat[Warping w/o EA]{\includegraphics[width=.245\textwidth]{img/ts-2-1.png}}
%     % \hfill
%     % \subfloat[Warping w/o EA]{\includegraphics[width=.245\textwidth]{img/ts-2-2.png}}
%     % \hfill
%     % \subfloat[Warping w/o EA]{\includegraphics[width=.245\textwidth]{img/ts-2-3.png}}
%     % \hfill
%     % \subfloat[Warping w/o EA]{\includegraphics[width=.245\textwidth]{img/ts-2-4.png}}
%     \captionof{figure}{Qualitative results of state-of-the-art stereo matching networks (StereoBase \cite{guo2023openstereo}) trained on different datasets under different adverse weather conditions. Both existing synthetic data and real data produce inconsistent and fragmented predictions. In contrast, models trained on the dataset generated by our proposed RobuSTereo generate more robust and accurate predictions.}
%     \label{fig: Tesear}
% \end{center}%
% }]

\begin{abstract}
    Learning-based stereo matching models struggle in adverse weather conditions due to the scarcity of corresponding training data and the challenges in extracting discriminative features from degraded images. These limitations significantly hinder zero-shot generalization to out-of-distribution weather conditions. In this paper, we propose \textbf{RobuSTereo}, a novel framework that enhances the zero-shot generalization of stereo matching models under adverse weather by addressing both data scarcity and feature extraction challenges. First, we introduce a diffusion-based simulation pipeline with a stereo consistency module, which generates high-quality stereo data tailored for adverse conditions. By training stereo matching models on our synthetic datasets, we reduce the domain gap between clean and degraded images, significantly improving the models’ robustness to unseen weather conditions. The stereo consistency module ensures structural alignment across synthesized image pairs, preserving geometric integrity and enhancing depth estimation accuracy. Second, we design a robust feature encoder that combines a specialized ConvNet with a denoising transformer to extract stable and reliable features from degraded images. The ConvNet captures fine-grained local structures, while the denoising transformer refines global representations, effectively mitigating the impact of noise, low visibility, and weather-induced distortions. This enables more accurate disparity estimation even under challenging visual conditions. Extensive experiments demonstrate that \textbf{RobuSTereo} significantly improves the robustness and generalization of stereo matching models across diverse adverse weather scenarios.
% We also provide a comprehensive evaluation of the generated datasets, highlighting significant improvements in stereo matching performance when our data is incorporated into existing stereo matching networks.
\end{abstract}

\section{Introduction}

Stereo matching is a fundamental task in computer vision that involves estimating the disparity between two input images to derive accurate depth information, which is crucial for various applications, including robotics~\cite{zhang2015building}, autonomous driving~\cite{orb2017slam2,guo2024camera}, and augmented reality~\cite{yang2019security}.

While existing methods~\cite{guo2023openstereo,xu2023iterative,guo2024stereo,cheng2025monster} perform well in most of the normal scenarios, they struggle in adverse weather conditions, such as rain and fog, which are unavoidable in real-world applications. Adverse weather conditions introduce outlier data that significantly degrade zero-shot stereo matching performance and severely impact downstream tasks like autonomous driving. A major factor contributing to this degradation is the substantial domain gap between data captured under normal conditions and in adverse environments. The degradation issue is further exacerbated in zero-shot scenarios, where models are directly applied to unseen weather conditions without fine-tuning. Moreover, images captured in adverse weather often suffer from low visibility, high noise, and strong mirror reflections, making it challenging for stereo matching models to extract reliable features and accurately estimate disparity.

\begin{figure}
    \centering
    \includegraphics[width=\linewidth]{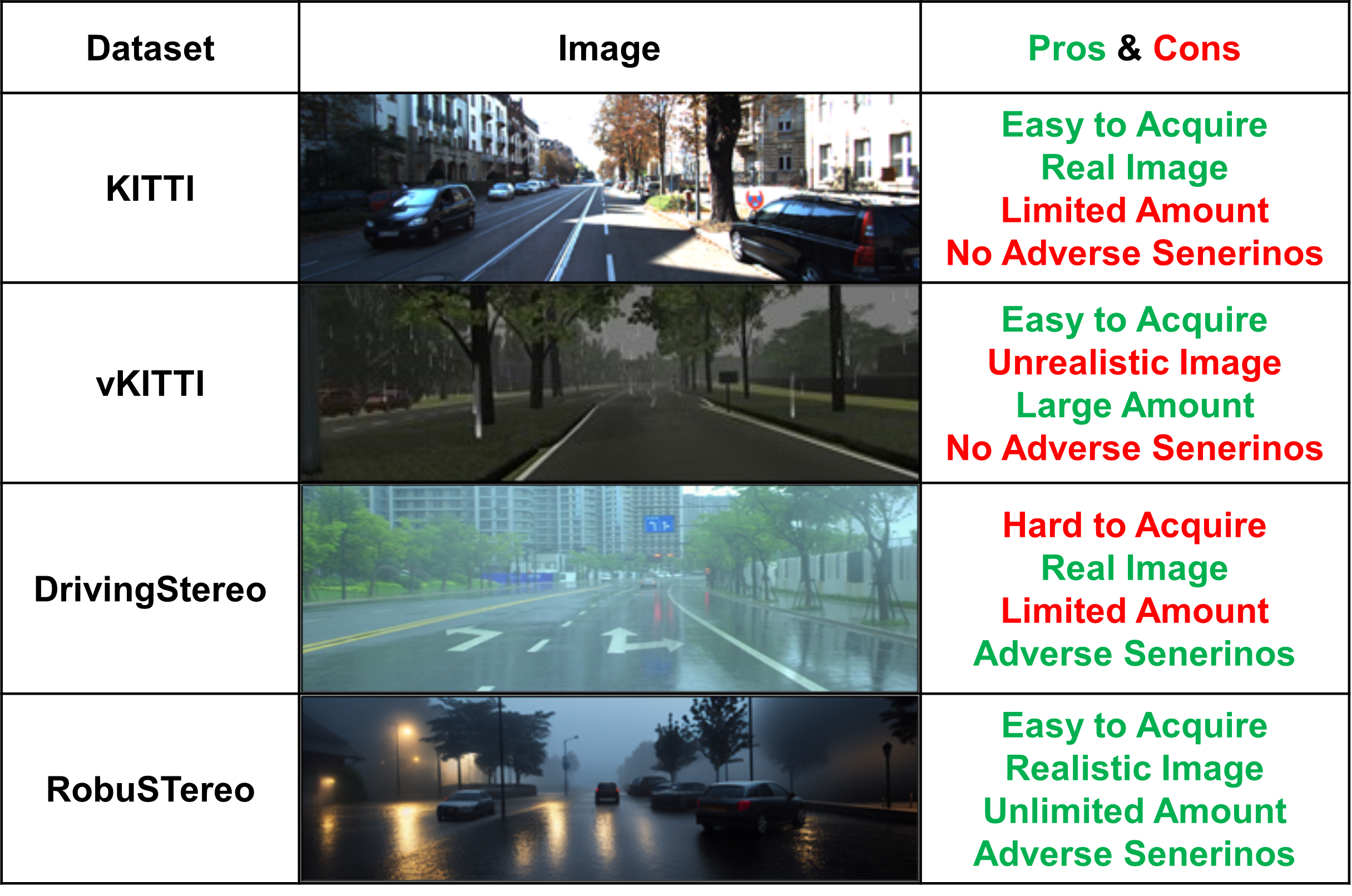}
        \caption{Comparison of different stereo datasets.}
    \label{fig:dataset-com}
    \vspace{-1em}
\end{figure}

Researchers attempt to address the domain gap problem by expanding datasets to include adverse weather conditions. As shown in Figure~\ref{fig:dataset-com}, some methods use graphics modeling~\cite{cabon2020vkitti2} or physics-based simulation~\cite{gaidon2016virtual} to generate stereo datasets. However, prior methods fail to capture complex light effects, such as strong mirror reflections on wet road surfaces, creating a domain gap between synthetic and real images. Others~\cite{el2024sid, bijelic2020seeing} try to collect real-world data in adverse weather, but disparity labels rely on depth sensors like LiDAR, Time-of-Flight (ToF), and Kinect, which perform poorly in bad weather, reducing dataset quality. Additionally, collecting real-world data is extremely time-consuming and labor-intensive. As a result, the scarcity of high-quality, large-scale training datasets under adverse weather conditions remains a significant challenge. 

A further limitation arises from the design of stereo matching encoders~\cite{guo2023openstereo,xu2023iterative,chang2018pyramid}. Networks pretrained on large image datasets under normal conditions struggle to extract stable and reliable features from low-quality images produced under adverse weather conditions, such as low visibility and high noise. Consequently, the extracted features tend to be significantly noisy, further reducing the accuracy of subsequent matching tasks and making it difficult for stereo matching models to achieve reliable zero-shot performance in such challenging environments. 

\begin{figure*}
\centering
\includegraphics[width=\linewidth]{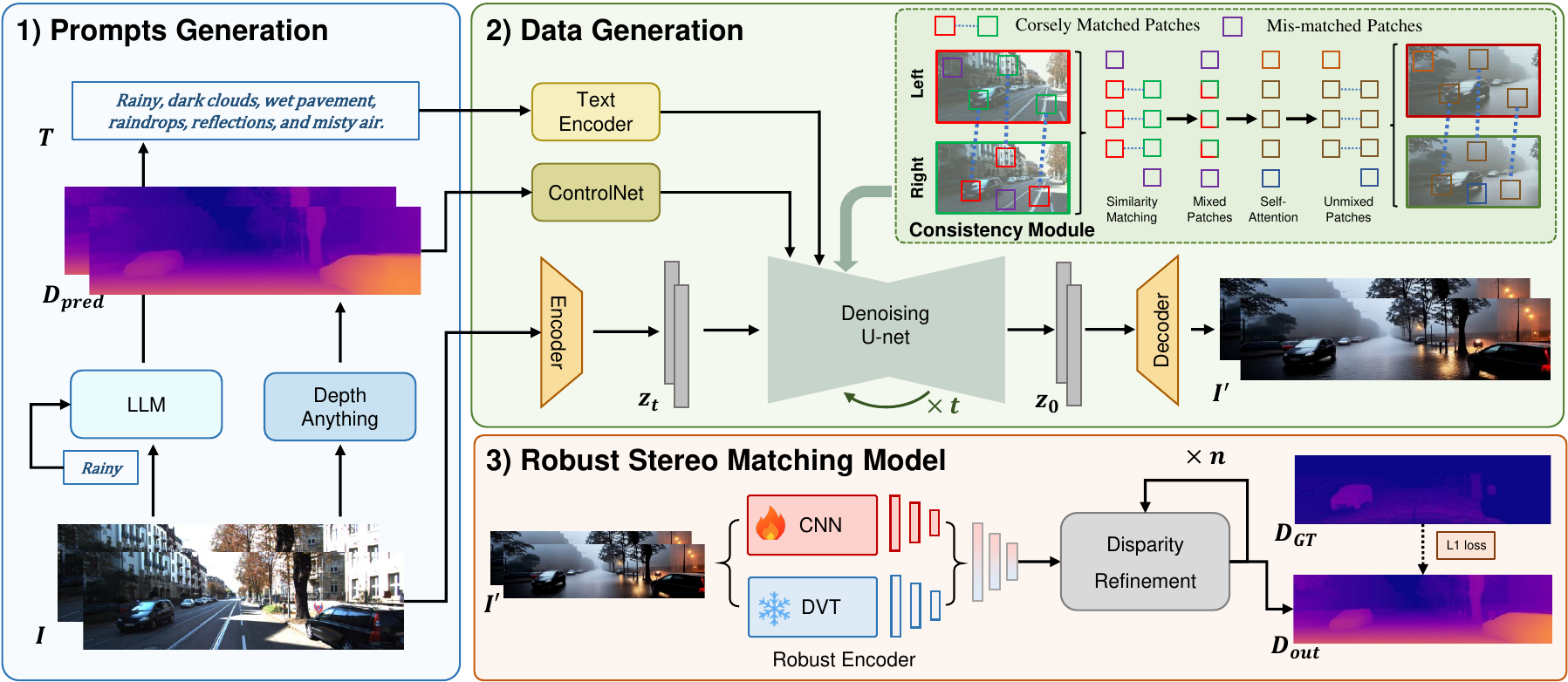}
\caption{RobuSTereo Framework for Stereo Matching under Adverse Weather Conditions. The framework consists of three components: (1) Prompts Generation, where depth and weather prompts are generated using a Large Language Model (LLM) and depth estimation to describe adverse weather conditions; (2) Data Generation, where the prompts and depth map guide a ControlNet and denoising U-Net to produce a weather-affected synthetic image; and (3) Robust Stereo Matching Model, where the synthetic image is used to train stereo matching networks with L1 loss for improved stereo matching under challenging weather.}
\label{fig:framework}
\vspace{-1em}
\end{figure*}

In this paper, we propose \textbf{RobuSTereo}, a novel stereo matching framework that improves zero-shot stereo matching performance under adverse weather conditions, shown in Figure~\ref{fig:framework}. On the one hand, we develop a diffusion-based method for generating stereo training data tailored to challenging weather scenarios. By utilizing a text-to-image diffusion model, our approach generates stereo images under adverse weather conditions using stereo datasets under normal scenarios and diverse text weather prompts. With powerful generative capabilities of diffusion models, we can produce large-scale, high-quality datasets for adverse weather conditions, boosting the generalization and performance of zero-shot stereo matching. Additionally, we incorporate a consistency module to ensure fine-grained consistency between stereo image pairs, which further improves the quality of synthesized training data. On the other hand, to further improve the feature extraction capacity, we propose a stereo matching network that integrates a robust feature encoder. The inclusion of robust feature encoder enhances the feature extraction capability of the network encoder, enabling it to capture more stable features in degraded imaging environments, which, in turn, improves both the accuracy and reliability of zero-shot stereo matching. Experimental results demonstrate that models trained on our synthetic dataset achieve superior zero-shot performance under challenging weather conditions.

To sum up, we make following contributions:

\begin{itemize}
    \item We propose \textbf{RobuSTereo}, a novel stereo matching framework that enhances zero-shot generalization under adverse weather by improving both data generation and feature extraction.
    \item We introduce a diffusion-based stereo data generation pipeline with a stereo consistency module, which synthesizes high-quality stereo pairs while preserving geometric alignment. This reduces the domain gap and enhances model robustness in adverse weather conditions.
    \item We design a robust feature encoder that integrates a specialized ConvNet and a denoising transformer, improving feature stability and stereo matching accuracy in degraded imaging conditions.
\end{itemize}

\section{Related Work}
\subsection{Stereo Matching Method}
Learning-based stereo matching methods have replaced traditional optimization techniques by leveraging convolutional neural networks (CNNs). The introduction of GCNet~\cite{kendall2017end} marked a significant shift by employing a 3D convolutional architecture to regularize a 4D cost volume. Building on this success, models such as PSMNet~\cite{chang2018pyramid}, GwcNet~\cite{guo2019group}, and GANet~\cite{zhang2019ga} have further improved accuracy. To enhance efficiency, cascade-based methods like CFNet~\cite{shen2021cfnet} were introduced. IGEV~\cite{xu2023iterative} proposed a novel module to capture non-local geometric and contextual information. Additionally, StereoBase~\cite{guo2023openstereo} provides a strong baseline model, further improving stereo network performance. Recently, StereoAnything~\cite{guo2024stereo}, pre-trained on both simulated and real datasets, achieves impressive zero-shot results. However, these methods are designed for normal environments and struggle to handle image degradation under harsh conditions. Therefore, improving model robustness for such challenging scenarios is essential.

\subsection{Datasets and Data Generation Method}
Stereo datasets can be broadly divided into real-world and synthetic datasets. Among real-world datasets, KITTI12~\cite{geiger2012we} and KITTI15~\cite{menze2015object} provide 200 training pairs of outdoor stereo images with sparse disparity labels from LiDAR measurements. Recently, large-scale datasets such as DIML~\cite{cho2021diml}, HRWSI~\cite{xian2020structure}, and IRS~\cite{wang2021irs} have emerged. While these datasets improve data volume and diversity, most are captured under normal lighting, lacking training data for complex scenes. Some researchers~\cite{el2024sid,bijelic2020seeing,yang2019drivingstereo} have collected data in adverse weather, but the quality is limited by the inaccuracy of LiDAR and other equipment in such conditions. Meanwile, synthetic datasets like SceneFlow~\cite{mayer2016large} and virtual KITTI(vKITTI)~\cite{cabon2020vkitti2} use computer graphics (CG) techniques to generate stereo images with precise disparity maps. vKITTI~\cite{cabon2020vkitti2} attempts to improve the diversity of the dataset by simulating different weather conditions such as rainy. However, CG-based methods struggle to replicate real-world complexity, leading to domain gaps that limit their real-world applicability.

\noindent\textbf{Image Diffusion.} Latent Diffusion Models (LDMs)~\cite{rombach2022high} have significantly reduced computational costs by applying denoising within the latent space. Conditioning methods include cross-attention mechanisms~\cite{avrahami2022blended,brooks2023instructpix2pix,gafni2024scene} and the conversion of segmentation masks~\cite{avrahami2023spatext} into tokens. Additionally, a variety of conditioning strategies~\cite{bar2023multidiffusion,bashkirova2023masksketch,zhang2023adding,li2024vidtome} have been developed to facilitate visual data generation based on diverse inputs, such as text, images, depth maps, videos and other forms of representations. Beyond image generation, diffusion models have also demonstrated exceptional capabilities in tasks like optical flow estimation~\cite{saxena2024surprising} and monocular depth estimation~\cite{ke2024repurposing,tosi2024diffusion,zhang2024atlantis}. Our method uses a diffusion generative model to generate a variety of stereo data pairs in complex environments to expand the diversity of stereo datasets under adverse weather.

\section{Method}

In this section, we first present the motivation of our proposed method. Next, we introduce data generation pipeline and the robust stereo matching network in detail. An illustration of the proposed framework is provided in Figure~\ref{fig:framework}.

\subsection{Motivation and Formulation}

% Accurate stereo matching in adverse weather is challenging, primarily due to the difficulty of collecting diverse datasets with precise disparity information. Existing datasets, such as DrivingStereo~\cite{yang2019drivingstereo}, are valuable but limited in diversity and scale. Moreover, obtaining reliable ground-truth disparity in conditions like rain, fog, and snow is challenging, as commonly used depth sensing devices, such as LiDAR and TOF, struggle to capture accurate depth information in such environments. In addition, most stereo matching networks are designed for normal environments. Their encoders lack the capacity to extract robust features from images with degraded quality caused by adverse weather.

Accurate zero-shot stereo matching in adverse weather, such as rain, fog, and snow, is not only essential but also highly challenging, as such conditions are unavoidable in real life and greatly reduce image quality, which directly affects stereo matching performance. The lack of large-scale, high-quality datasets for adverse weather further worsens these challenges, making it difficult to achieve reliable stereo matching. Existing datasets, such as DrivingStereo~\cite{yang2019drivingstereo}, are limited in diversity and scale, and sensors like LiDAR struggle to provide accurate disparity under such conditions. Furthermore, most stereo matching networks lack robust encoders capable of handling the degraded image quality caused by adverse weather.

To address these challenges, we propose \textbf{RobuSTereo}, which generates realistic synthetic stereo data under various adverse weather conditions and introduces a more powerful stereo matching encoder. Our method enables scalable data generation with accurate disparity maps, offering greater diversity and ease of acquisition than existing datasets. Furthermore, we optimize stereo matching networks by introducing a robust feature encoder, enhancing feature extraction stability and improving performance in complex adverse environments.

% Therefore, we propose RobuSTereo, which addresses these challenges by generating realistic synthetic stereo data across various adverse weather conditions and introducing a more powerful stereo matching network to handle image quality degradation. Using images captured under normal weather conditions and textual prompts, our approach enables extensive data generation with accurate disparity maps. As a result, our dataset offers advantages in simplicity of acquisition, scene diversity, and scalability, representing a significant improvement over existing datasets and stereo image synthesis techniques for challenging environments. Furthermore, to address the issue of poor image feature extraction under adverse weather conditions, we optimized the encoder of traditional stereo matching networks. Specifically, we introduced an encoder structure integrated with Denoising-ViT, enhancing the stability and capability of the model's encoder and the model's performance in complex and challenging environments.

\subsection{Stereo Data Generation for Adverse Weather}

A training pair $(\mathbf{I}_R, \mathbf{I}_L, \mathbf{D})$ of a stereo dataset consists of two parts: stereo images $\mathbf{I}_R, \mathbf{I}_L \in \mathbb{R}^{3\times H\times W}$ and the disparity map $\mathbf{D}\in\mathbb{R}^{ H\times W}$ (typically the disparity of the left view). Most of the existing image datasets are data under normal weather domain $(\mathbf{I}_R, \mathbf{I}_L, \mathbf{D})_{norm}$. We hope to transfer them to adverse environments domain $(\mathbf{I}_R', \mathbf{I}_L', \mathbf{D})_{adv}$ through style transfer, where $\mathbf{I}'$ represents image under adverse weather. Then we use these data pairs to train a more robust stereo matching network. Therefore, the task can be formulated as $(\mathbf{I}_R', \mathbf{I}_L', \mathbf{D}) = \mathcal{T}(\mathbf{I}_R, \mathbf{I}_L, \mathbf{D})$, where $\mathcal{T}$ stands for the desired transformation method. 

\noindent\textbf{Diffusion-based Stereo Image Generation.} 
IDMs has obvious advantages in image style transfer and image editing. Therefore, we introduce the diffusion method using ControlNet to achieve stereo image generation.

First, we generate descriptors for the target domain. With the powerful understanding and generalization capabilities of the large language models (LLMs), we use them to help us generate powerful prompt keywords. Taking the conversion of an image pair to a rainy day as an example, we input the source image into LLMs~\cite{achiam2023gpt} and ask it to generate prompt words under the corresponding conditions according to the image content, like \textit{describe a rainy day of this image in a few keywords}. We collect the prompts $T$ \textit{ = Rainy, dark clouds, wet pavement, raindrops, reflections, and misty air} for subsequent image style transfer.

\begin{figure}
\centering
\includegraphics[width=\linewidth]{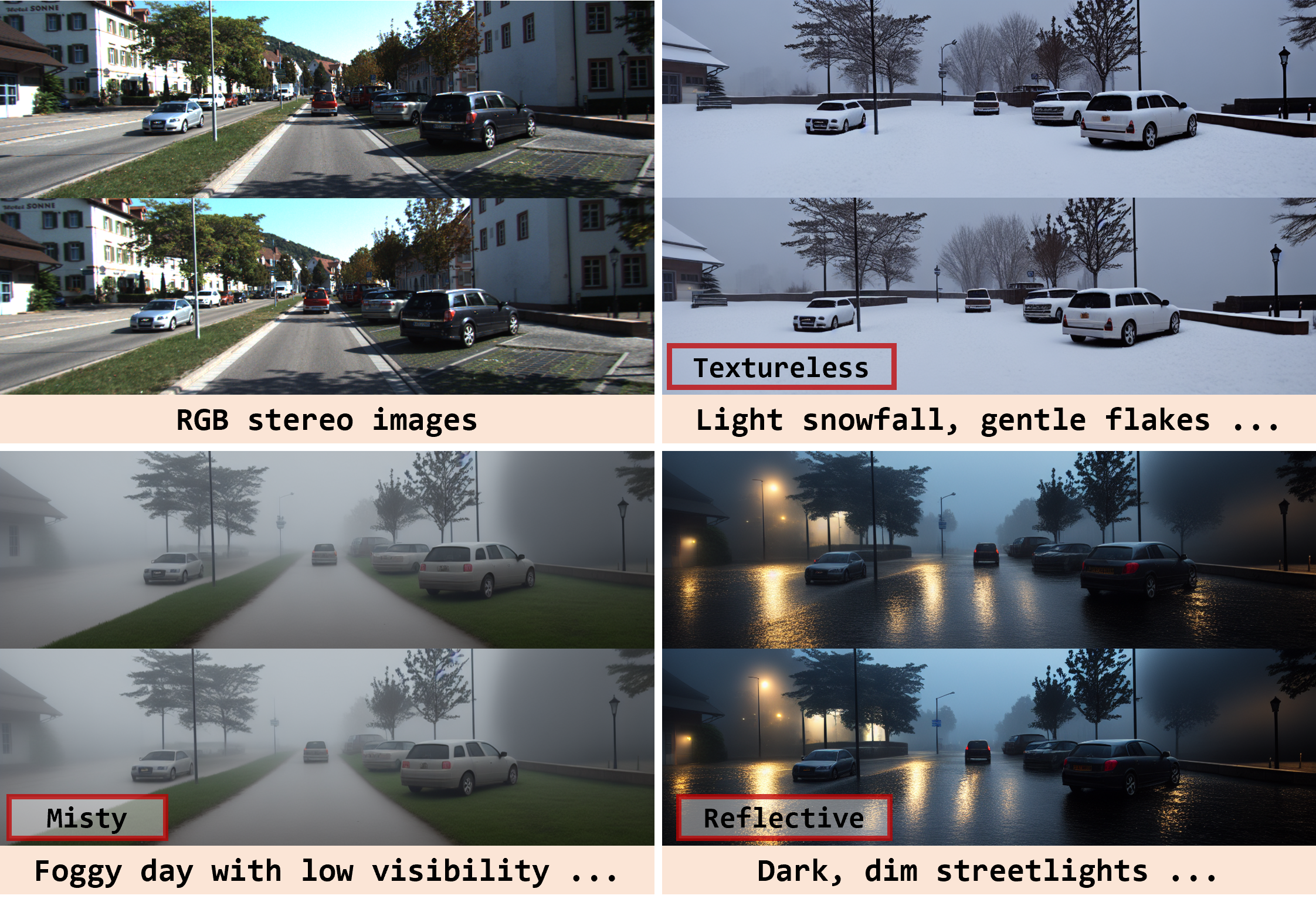}
\caption{Visualization of generated data under adverse weather using our proposed RobuSTereo from stereo images under normal conditions. From top to bottom: RGB stereo images, Snowy, Foggy and Rainy.}
\label{fig:generated-example}
\vspace{-1em}
\end{figure}

Then, we use Depth2Image ControlNet~\cite{zhang2023adding} as a guidance for image content to ensure that the generated image is consistent with the original image in depth information. In this way, the generated image can match the original disparity GT ($\mathbf{D}_{GT}$). Considering that $\mathbf{D}_{GT}$ of some stereo datasets is sparse and cannot be used as the input of the controlnet, we first use the depth estimation network~\cite{depth_anything_v2} to predict the depth map $\mathbf{D}_{pred}$ of the image and use it as the control input of the controlnet. Since the existing images are generally image pairs in a clear and simple environment, the image depth obtained in this step has a high credibility. The conditioning process is:
\begin{align}
    &\mathbf{D}_{pred} = (\mathcal{D}(\mathbf{I}_R), \mathcal{D}(\mathbf{I}_L)), \\
    &\mathbf{c} = \mathcal{F}_{CtrlNet}(\mathbf{z}_t, \mathbf{D}_{pred}, T),
\end{align}
where $\mathcal{F}_{CtrlNet}(\cdot)$ represents Depth2Image ControlNet~\cite{zhang2023adding}, $\mathcal{D}(\cdot)$ is DepthAnythingV2~\cite{depth_anything_v2}, $T$ is relevant text prompts, $t$ denotes the $t$-th step of the backward diffusion process and $c$ is conditioning feature, which $\mathbb{c}$ is then utilized in the Stable Diffusion (SD) generation process
\begin{align}
    \mathbf{I'} = \mathcal{F}_{SD}(\mathbf{z}_t, T|\mathbf{c}),
\end{align}
$\mathbf{I'}$ is for the generated image for target domain, and $\mathcal{F}_{SD}(\cdot)$ stands for the generation process of pretrained Stable Diffusion model conditioned by a ControlNet. Diffusion-base methodology enables the generation of a wide range of stereo image pairs that maintain the underlying scene structure while exhibiting varied visual characteristics, simulating different adverse weather conditions. Example sythetic images are shown in Figure~\ref{fig:generated-example}. Theoretically, we can generate an unlimited amount of relevant data.

\noindent\textbf{Coherence-Enhanced Consistency Module.} Diffusion-based methods have clear advantages in generating detailed and diverse content. However, when generating stereo images, excessive diversity in the generated content can result in the left and right images not corresponding properly. Such inconsistencies make generated images unsuitable for training stereo networks. To address this issue, we introduce specific consistency constraints during the generation process to enhance the content similarity between the left and right images. This ensures that the generated stereo images are more aligned and suitable for stereo network training.

Inspired by video editing methods, we propose a Disparity Fusion Method (DFM) to improve the consistency of generated images. Given the input image patches $\mathbf{P}$, DFM first divides the patches into a source set (\textit{src}) and a destination set (\textit{dst}), and computes the patch-wise similarity between these two sets. The similarity between two patches is calculated based on their disparity similarity and image similarity. Each patch in the \textit{src} set is linked to its most similar patch in the \textit{dst} set, and the top-\textit{n} most similar patch pairs are fused. The formulation is expressed as 
\begin{align}
&\mathbf{E} = Match(src, dst, n), \\
&\mathbf{P}_m = \mathcal{M}(\mathbf{P}, \mathbf{E}), \quad \mathbf{P}_u = \mathcal{U}(\mathbf{P}_m, \mathbf{E}),
\end{align}
$Match(\cdot)$ represents the process of finding the top-\textit{n} most similar patch pairs, while $\mathcal{M}(\cdot)$ and $\mathcal{U}(\cdot)$ denote the patch fusion and un-fusion processes, respectively, based on the matching map $\mathbf{E}$.
In the context of diffusion-based stereo image generation, the input and output patches are denoted as $\mathbf{P}_{in}, \mathbf{P}_{out} \in \mathbb{R}^{2\times N\times C}$, where $N$ is the number of patches, and $C$ is the feature dimension. Based on Equation (4), we perform the fusion process as
\begin{align}
&\mathbf{P}_{m} = \mathcal{M}(\mathbf{P}_{in}, Match(\mathbf{P}_{in}^{left}, \mathbf{P}_{in}^{right}, n)), \\
&\mathbf{P}_{out} = \mathcal{U}(Attn(\mathbf{P}_{m}), Match(\mathbf{P}_{in}^{left}, \mathbf{P}_{in}^{right}, n)).
\end{align}
The fused patches $\mathbf{P}_{m}$ are then passed through a self-attention mechanism, denoted as $Attn(\cdot)$, to identify and refine consistent features. This approach significantly improves the multi-view consistency of the generated stereo images, ensuring that the left and right views are well-aligned and suitable for stereo network training.

\begin{table*}[ht!]
\centering
\small
\begin{tabular}{lccccccccccc}
\toprule
\multirow{2}{*}{Networks}  & \multirow{2}{*}{Publication} & \multicolumn{2}{c}{\textbf{Rainy}} & \multicolumn{2}{c}{\textbf{Sunny}} & \multicolumn{2}{c}{\textbf{Foggy}} & \multicolumn{2}{c}{\textbf{Cloudy}} & \multicolumn{2}{c}{\textbf{Overall}}\\
\cmidrule(lr){3-4} \cmidrule(lr){5-6} \cmidrule(lr){7-8} \cmidrule(lr){9-10} \cmidrule(lr){11-12}
& & EPE $\downarrow$ & D1 $\downarrow$ &  EPE $\downarrow$ & D1 $\downarrow$ &  EPE $\downarrow$ & D1 $\downarrow$ & EPE $\downarrow$ & D1 $\downarrow$ &  EPE $\downarrow$ & D1 $\downarrow$\\
\midrule
PSMNet~\cite{chang2018pyramid}&CVPR'18&20.864&50.862&3.668&27.499&19.562&58.040&4.437&30.989&12.133&41.848 \\
CFNet~\cite{shen2021cfnet}&CVPR'21&4.211&23.558&2.178&15.060&3.442&25.910&3.392&23.277&3.306&21.951 \\
GwcNet~\cite{guo2019group}&CVPR'19&6.211&48.851&2.957&23.896&4.721&43.887&3.760&29.950&4.412&36.646 \\
COEX~\cite{bangunharcana2021correlate}&IROS'21&4.531&28.959&2.490&17.264&3.093&22.694&2.995&21.849&3.277&22.692\\
FADNet++~\cite{wang2020fadnet}&ICRA'20&2.852&24.234&1.999&15.019&2.214&19.566&2.063&15.985&2.282&18.701 \\
CasStereo~\cite{Gu_2020_CVPR_CasStereo}&CVPR'20&5.013&33.692&3.612&22.732&4.143&31.444&3.863&26.123&4.158&28.498\\
IGEV~\cite{xu2023iterative}&CVPR'23&1.879&10.955&1.215&5.080&1.253&6.582&1.077&4.196&1.356&6.704 \\
StereoBase~\cite{guo2023openstereo}&ArXiv'23&1.695&8.610&1.198&5.023&1.224&5.980&1.090&4.281&1.302&5.974 \\
NMRF~\cite{guan2024neural}&CVPR'24&2.490&12.073&1.046&3.674&\underline{1.089}&\underline{4.337}&0.976&3.162&1.401&5.811\\
Selective-IGEV~\cite{wang2024selective}&CVPR'24&1.178&5.397&1.103&4.296&2.172&13.660&1.127&4.820&1.395&7.043\\
StereoAnything~\cite{guo2024stereo}&ArXiv'24&1.144&5.395&\underline{0.984}&\underline{2.972}&1.134&4.821&\underline{0.906}&\underline{2.273}&\underline{1.042}&\underline{3.865}\\
LightStereo~\cite{guo2024lightstereo}&ICRA'25&\underline{1.105}&\underline{4.846}&1.078&3.609&1.155&4.927&1.014&3.046&1.088&4.107\\
MonSter~\cite{cheng2025monster}&CVPR'25&1.153&5.335&1.030&3.509&1.152&5.275&0.987&3.180&1.081&4.325\\
% \rowcolor{graycolor}StereoBase-Ours&&1.003&2.221&0.834&1.834&0.867&2.383&0.797&1.764&0.875&2.050  \\
\rowcolor{graycolor}  RobuSTereo (Ours) &-&\textbf{0.973}&\textbf{1.939}&\textbf{0.785}&\textbf{1.493}&\textbf{0.853}&\textbf{1.610}&\textbf{0.735}&\textbf{1.349}&\textbf{0.836}&\textbf{1.598}\\
\bottomrule
\end{tabular}
\caption{Comparison of model performance on DrivingStereo~\cite{yang2019drivingstereo} with stereo matching models. ``Ours" refers to our model trained on our generated dataset. All models are validated on the DrivingStereo weather subset. Most methods are trained on SceneFlow~\cite{mayer2016large}, except StereoAnything~\cite{guo2024stereo}, LightStereo~\cite{guo2024lightstereo}, and MonSter~\cite{cheng2025monster}, which are trained on mixed datasets combining synthetic and real-world data.}
\label{tab-SM-DS}
\vspace{-1em}
\end{table*}

\subsection{Stereo Networks with Robust Feature Encoder}

In adverse weather, RGB image imaging is often affected by noise and low visibility, which poses significant challenges for stereo matching. Traditional stereo matching methods typically rely on simple encoders pre-trained on ImageNet~\cite{imagenet_cvpr09} or other large-scale RGB datasets as their feature extraction modules. However, these encoders are not specifically designed to handle the degraded image quality encountered in adverse weather conditions, making them difficult to process such challenging scenes.

In this paper, in order to improve the stability of feature extraction by the encoder, we propose a robust feature encoder by combining Denosing-Vision Transformer (DVT)~\cite{yang2024dvt} and VGG19~\cite{vgg}. DVT captures richer and more robust high-dimensional feature representations, including both semantic and high-level information, while VGG focuses on extracting detailed pixel-wise features. Compared to traditional CNN-based encoder or ViT encoder, DVT can reduce the noise of features at the feature level to obtain more stable features, solving the problem of poor image quality in complex weather conditions. 

Formally, given a pair of left and right images, $\mathbf{I}_L, \mathbf{I}_R \in \mathbb{R}^{H \times W \times 3}$, we first utilize the well-established VGG19~\cite{vgg} to extract feature. VGG19~\cite{vgg} generates multi-scale pyramid features, producing detailed feature maps at different resolutions: $f_c^{(i)} \in \mathbb{R}^{C_i \times \frac{H}{i} \times \frac{W}{i}}$, where $i \in \{4, 8, 16\}$ represents the scaling factor. These multi-scale features provide a rich hierarchical representation of the input images. To further enhance the feature extraction process, we incorporate DVT, which generates an additional high-dimensional and robust feature map, $f_c^{(32)} \in \mathbb{R}^{C_{32} \times \frac{H}{32} \times \frac{W}{32}}$, capturing high-level semantic and contextual information, which plays a critical role in improving the robustness and accuracy of stereo matching, especially in complex scenarios.

% Formally, given a pair of left and right images, $\mathbf{I}_L, \mathbf{I}_R \in \mathbb{R}^{H \times W \times 3}$, we first utilize the well-established VGG19~\cite{vgg} to extract features. VGG19~\cite{vgg} generates multi-scale pyramid features, producing detailed feature maps at multiple resolutions: $f_c^{(i)} \in \mathbb{R}^{C_i \times \frac{H}{i} \times \frac{W}{i}}$, where $i \in \{4, 8, 16\}$ represents the scaling factor. These multi-scale features provide a rich hierarchical representation of the input images, effectively capturing both fine details and structural information.

% To further enhance the feature extraction process, we incorporate the Dynamic Vision Transformer (DVT), which generates an additional high-dimensional and robust feature map, $f_c^{(32)} \in \mathbb{R}^{C_{32} \times \frac{H}{32} \times \frac{W}{32}}$. This feature map captures high-level semantic and contextual information, which plays a critical role in improving the robustness and accuracy of stereo matching, especially in complex scenarios.

Finally, the disparity refinement network is designed following on the method proposed in~\cite{guo2023openstereo}, which enhances depth estimation by refining the initial disparity maps iteratively. The integration of advanced feature extraction improves the prediction accuracy under adverse weather.

\section{Experiments}
In this section, we introduce the datasets, implementation details, and conduct zero-shot comparisons across different datasets. Ablation studies validate the effectiveness of our main components, with additional analysis on visual SLAM provided in the supplementary materials.

\begin{table*}[t]
\small
\centering
\begin{tabular}{llcccccccccc}
\toprule
\multirow{2}{*}{Networks} & \multirow{2}{*}{Datasets} & \multicolumn{2}{c}{\textbf{Snow}} & \multicolumn{2}{c}{\textbf{Rain}} & \multicolumn{2}{c}{\textbf{Dense Fog}} & \multicolumn{2}{c}{\textbf{Light Fog}} & \multicolumn{2}{c}{\textbf{Overall}}\\
\cmidrule(lr){3-4} \cmidrule(lr){5-6} \cmidrule(lr){7-8} \cmidrule(lr){9-10} \cmidrule(lr){11-12} 
& & EPE $\downarrow$ & D1 $\downarrow$ &  EPE $\downarrow$ & D1 $\downarrow$ &  EPE $\downarrow$ & D1 $\downarrow$ & EPE $\downarrow$ & D1 $\downarrow$ &  EPE $\downarrow$ & D1 $\downarrow$\\
\midrule
% Selective-IGEV~\cite{wang2024selective}&SceneFlow \\
StereoAnything~\cite{guo2024stereo}&MIX&4.265&28.734&3.440&21.955&6.055&27.290&2.973&16.170&4.204&26.526\\
LightStereo~\cite{guo2024lightstereo}&MIX&\underline{3.894}&30.104&\underline{3.034}&24.836&5.904&31.083&2.622&19.118&\underline{3.853}&28.431\\
\midrule
\multirow{4}{*}{StereoBase~\cite{guo2023openstereo}} 
& SceneFlow~\cite{mayer2016large}&6.567&33.623&5.150&28.592&6.347&30.958&3.071&20.955&6.007&31.434\\
& KITTI~\cite{menze2015object}&8.001&29.802&5.918&21.333&\underline{5.533}&22.101&2.801&13.634&6.976&26.449\\
& vKITTI~\cite{cabon2020vkitti2}&5.519&34.634&4.447&28.408&6.621&33.801&3.264&22.450&5.259&32.573\\
\rowcolor{graycolor} \cellcolor{white}&RST-Dataset&7.059&\underline{27.532}&4.516&\underline{18.107}&5.601&\underline{20.429}&\underline{2.438}&\underline{10.301}&6.158&\underline{24.026}\\
\midrule
\rowcolor{graycolor}RobuSTereo (Ours)&RST-Dataset&\textbf{3.409}&\textbf{23.438}&\textbf{2.577}&\textbf{15.656}&\textbf{5.317}&\textbf{20.128}&\textbf{2.120}&\textbf{10.074}&\textbf{3.359}&\textbf{20.881}  \\
\bottomrule
\end{tabular}
\caption{Comparison of zero-shot performance on SeeingThroughFog~\cite{bijelic2020seeing}. StereoAnything~\cite{guo2024stereo}, LightStereo~\cite{guo2024lightstereo} are trained on mixed synthetic and real-world datasets. Stereobase~\cite{guo2023openstereo} model are trained on four different datasets.}
\label{tab-SM-STF}
% \vspace{-0.3em}
\end{table*}

\begin{figure*}[]
\centering
\small
\setlength\tabcolsep{1pt}
\begin{tabular}{ccccc}
\includegraphics[width=0.195\textwidth]{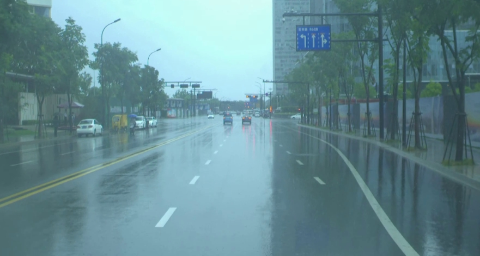} & 
\includegraphics[width=0.195\textwidth]{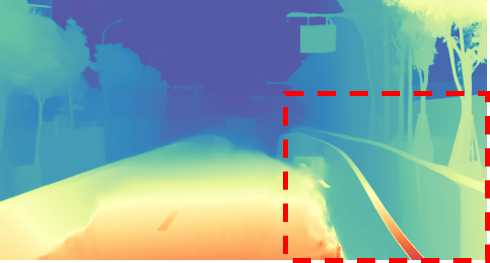} & 
\includegraphics[width=0.195\textwidth]{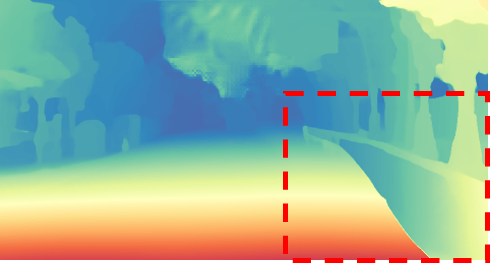} & 
\includegraphics[width=0.195\textwidth]{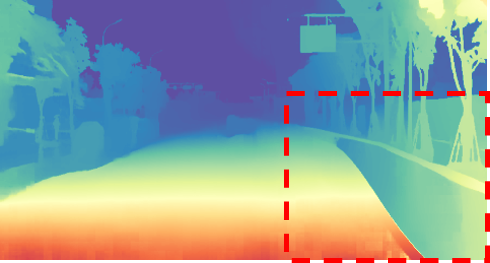} & 
\includegraphics[width=0.195\textwidth]{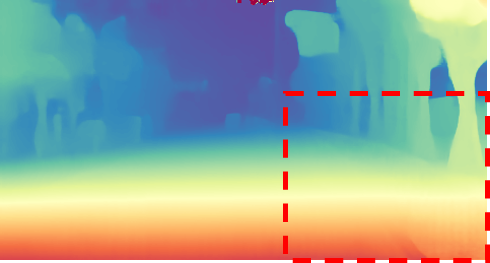} \\
\includegraphics[width=0.195\textwidth]{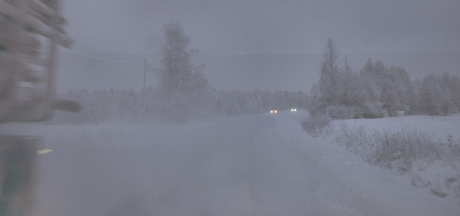} & 
\includegraphics[width=0.195\textwidth]{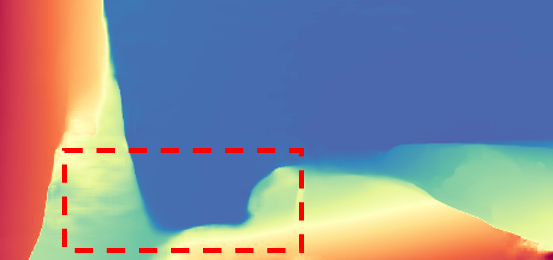} & 
\includegraphics[width=0.195\textwidth]{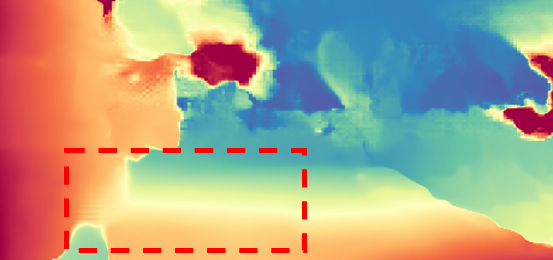} & 
\includegraphics[width=0.195\textwidth]{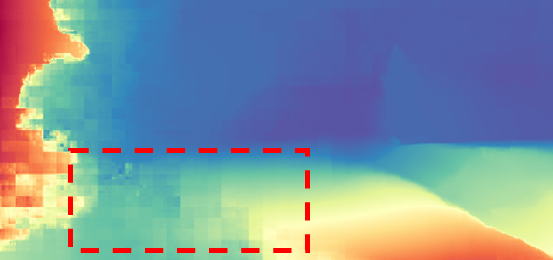} & 
\includegraphics[width=0.195\textwidth]{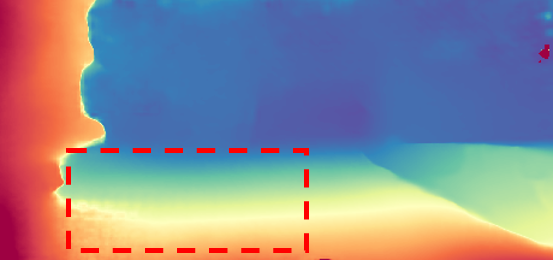} \\
\includegraphics[width=0.195\textwidth]{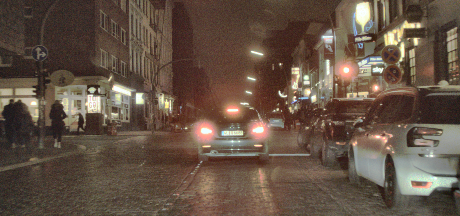} & 
\includegraphics[width=0.195\textwidth]{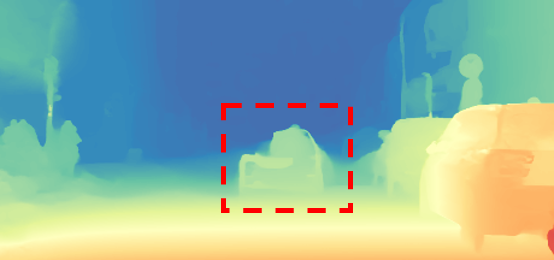} & 
\includegraphics[width=0.195\textwidth]{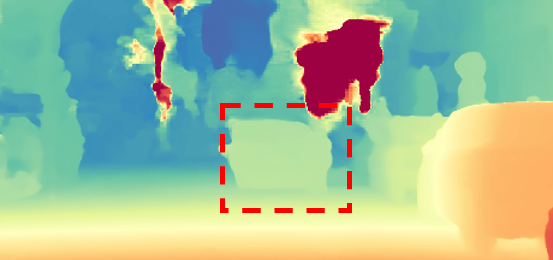} & 
\includegraphics[width=0.195\textwidth]{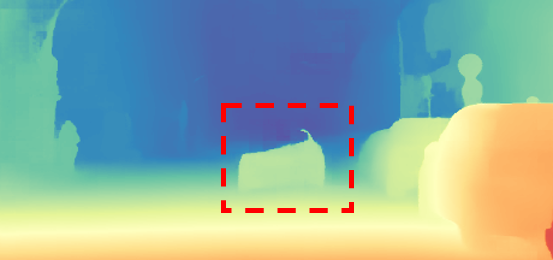} & 
\includegraphics[width=0.195\textwidth]{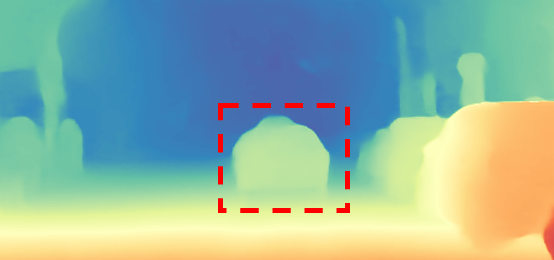} \\
\includegraphics[width=0.195\textwidth]{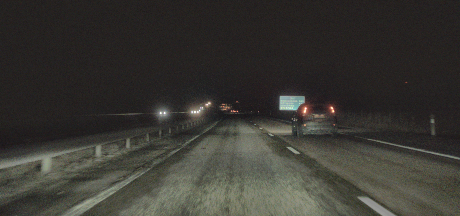} & 
\includegraphics[width=0.195\textwidth]{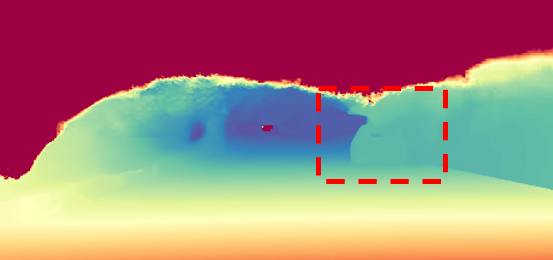} & 
\includegraphics[width=0.195\textwidth]{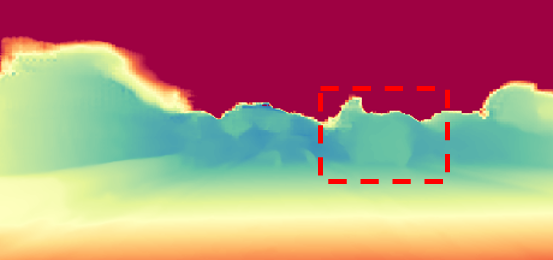} & 
\includegraphics[width=0.195\textwidth]{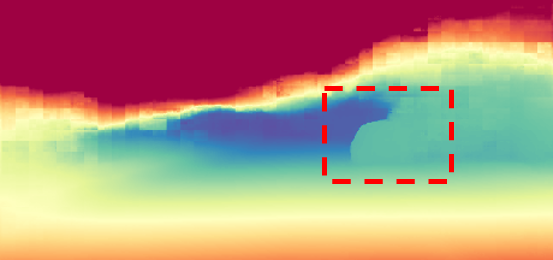} & 
\includegraphics[width=0.195\textwidth]{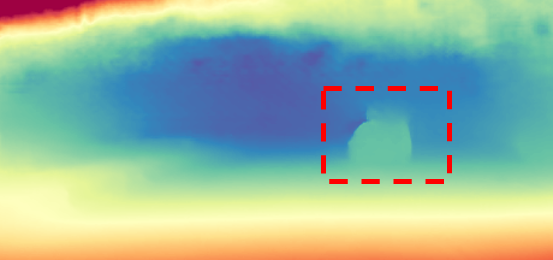} \\
\includegraphics[width=0.195\textwidth]{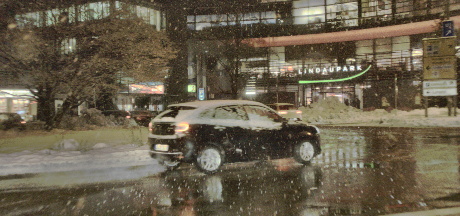} & 
\includegraphics[width=0.195\textwidth]{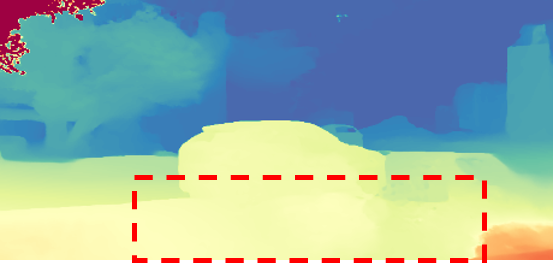} & 
\includegraphics[width=0.195\textwidth]{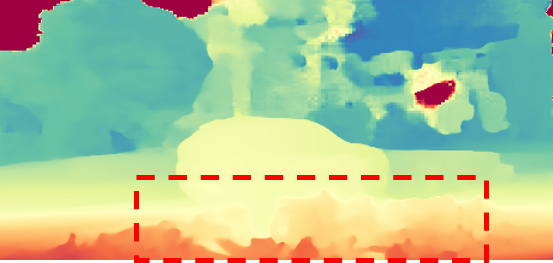} & 
\includegraphics[width=0.195\textwidth]{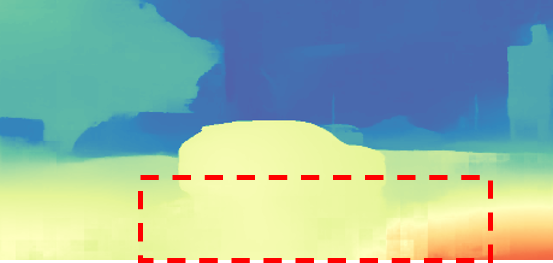} & 
\includegraphics[width=0.195\textwidth]{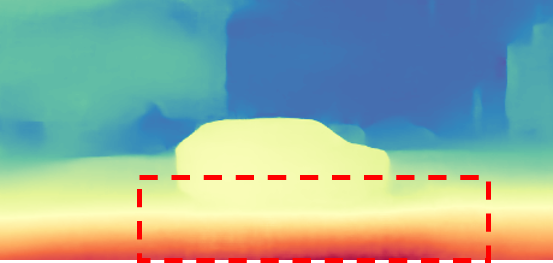} \\
Left Image & StereoBase-SF~\cite{guo2023openstereo} & StereoBase-KITTI~\cite{guo2023openstereo} & StereoAnything~\cite{guo2024stereo} & Ours
\end{tabular}
\caption{Qualitative results under adverse conditions. Our method produces more accurate and consistent disparity maps, preserving fine details and reducing artifacts compared to other approaches, as highlighted in the \textcolor{red}{red} boxes.}
\label{fig:vis-dataset}
% \vspace{-0.7em}
\end{figure*}

\subsection{Experimental Setup}

\noindent\textbf{Datasets.} In our experiments, we utilize several datasets for training and evaluation. \textbf{SceneFlow}~\cite{mayer2016large}, \textbf{KITTI 2012}~\cite{geiger2012we}, \textbf{KITTI 2015}~\cite{menze2015object}, and \textbf{Virtual KITTI V2} (\textbf{vKITTI})~\cite{cabon2020vkitti2} is used for training. \textbf{DrivingStereo}~\cite{yang2019drivingstereo} and \textbf{SeeingThroughFog}~\cite{bijelic2020seeing} serve as test sets for performance evaluation under different adverse weather conditions for both quantitative evaluations and visualization. We provide more detailed dataset information in the supplementary material.

\begin{table*}[t]
\small
\centering
\begin{tabular}{llcccccccccc}
\toprule
\multirow{2}{*}{Networks} & \multirow{2}{*}{Datasets} & \multicolumn{2}{c}{Rainy} & \multicolumn{2}{c}{Sunny} & \multicolumn{2}{c}{Foggy} & \multicolumn{2}{c}{Cloudy} & \multicolumn{2}{c}{Overall}\\
\cmidrule(lr){3-4} \cmidrule(lr){5-6} \cmidrule(lr){7-8} \cmidrule(lr){9-10} \cmidrule(lr){11-12} 
& & EPE $\downarrow$ & D1 $\downarrow$ &  EPE $\downarrow$ & D1 $\downarrow$ &  EPE $\downarrow$ & D1 $\downarrow$ & EPE $\downarrow$ & D1 $\downarrow$ &  EPE $\downarrow$ & D1 $\downarrow$\\
\midrule
\multirow{4}{*}{PSMNet~\cite{chang2018pyramid}} & SceneFlow~\cite{mayer2016large}&20.864&50.862&3.668&27.499&19.562&58.040&4.437&30.989&12.133&41.848\\
& KITTI~\cite{menze2015object}  
&1.632&\underline{6.077}&\underline{1.241}&\underline{3.687}&\underline{1.208}&\underline{5.107}&\underline{0.973}&\underline{2.648}&\underline{1.264}&\underline{4.380} \\
& vKITTI~\cite{cabon2020vkitti2} 
&\underline{1.579}&9.007&1.418&6.404&1.342&8.439&1.036&3.463&1.344&6.829  \\
\rowcolor{graycolor} \cellcolor{white}& RST-Dataset (Ours)  &\textbf{1.208}&\textbf{4.723}&\textbf{1.058}&\textbf{3.317}&\textbf{1.076}&\textbf{4.722}&\textbf{0.912}&\textbf{2.367}&\textbf{1.064}&\textbf{3.783}\\
\midrule
\multirow{4}{*}{IGEV~\cite{xu2023iterative}} & SceneFlow~\cite{mayer2016large} &1.879&10.955&1.215&5.080&1.253&6.582&1.077&4.196&1.356&6.704  \\
& KITTI~\cite{menze2015object}  &\underline{1.456}&\underline{4.915}&\textbf{0.838}&\textbf{2.026}&\underline{0.917}&\textbf{1.988}&\textbf{0.822}&\textbf{2.046}&\underline{1.009}&\underline{2.792}  \\
& vKITTI~\cite{cabon2020vkitti2}  &1.533&7.529&1.151&3.679&1.208&5.471&0.992&3.038&1.221&4.930  \\
\rowcolor{graycolor} \cellcolor{white}& RST-Dataset (Ours) &\textbf{1.082}&\textbf{3.467}&\underline{0.920}&\underline{2.308}&\textbf{0.902}&\underline{2.184}&\underline{0.867}&\underline{2.140}&\textbf{0.943}&\textbf{2.525}\\\midrule
\multirow{4}{*}{StereoBase~\cite{guo2023openstereo}} 
& SceneFlow~\cite{mayer2016large} &1.695&8.610&1.198&5.023&1.224&5.980&1.090&4.281&1.302&5.974\\
& KITTI~\cite{menze2015object}  &\underline{1.150}&\underline{2.857}&\underline{0.840}&\underline{1.947}&\underline{0.920}&\underline{2.408}&\underline{0.820}&\underline{1.904}&\underline{0.927}&\underline{2.279}  \\
& vKITTI~\cite{cabon2020vkitti2} &1.244&6.949&1.108&4.005&1.288&6.399&0.992&3.273&1.159&5.157   \\
\rowcolor{graycolor} \cellcolor{white}&  RST-Dataset (Ours) &\textbf{1.003}&\textbf{2.221}&\textbf{0.834}&\textbf{1.834}&\textbf{0.867}&\textbf{2.383}&\textbf{0.797}&\textbf{1.764}&\textbf{0.875}&\textbf{2.050} \\
\bottomrule
\end{tabular}
\caption{Comparison of zero-shot performance on DrivingStereo~\cite{yang2019drivingstereo} dataset using different datasets. We train three representative models, including PSMNet~\cite{chang2018pyramid}, IGEV~\cite{xu2023iterative}, and StereoBase~\cite{guo2023openstereo}. Among them, StereoBase represents the state-of-the-art model.}
\label{tab-DATA-DS}
\end{table*}

% \subsection{Implementation Details}

\noindent\textbf{Image Generation Configurations.} We use the Diffusers library to modify and deploy Stable Diffusion~\cite{rombach2022high} (SD) and ControlNet~\cite{zhang2023adding}. SD (version 1.5) acts as the image generator, with a DDIM scheduler and 50 sampling steps. For depth-conditioned generation, we use ControlNet~\cite{zhang2023adding} (version \textit{control\_v11f1p\_sd15\_depth}) and Depth-Anything V2~\cite{depth_anything_v2} for monocular depth estimation. Source images from KITTI~\cite{geiger2012we,menze2015object} and vKITTI~\cite{cabon2020vkitti2} are used to generate stereo data for adverse weather. The synthetic dataset  is named Robust-Dataset (\textbf{RST-Dataset}).

\noindent\textbf{Stereo Matching Models.} To evaluate our stereo matching framework, we compare it against both traditional and state-of-the-art methods, including PSMNet~\cite{chang2018pyramid}, CFNet~\cite{shen2021cfnet}, GwcNet~\cite{guo2019group}, COEX~\cite{bangunharcana2021correlate}, FADNet++~\cite{wang2020fadnet}, CasStereo~\cite{Gu_2020_CVPR_CasStereo}, IGEV~\cite{xu2023iterative}, StereoBase~\cite{guo2023openstereo}, NMRF~\cite{guan2024neural}, Selective-IGEV~\cite{wang2024selective}, LightStereo~\cite{guo2024lightstereo}, StereoAnything~\cite{guo2024stereo} and MonSter~\cite{cheng2025monster}. For dataset-level experiments, we focus on PSMNet~\cite{chang2018pyramid}, IGEV~\cite{xu2023iterative}, and StereoBase~\cite{guo2023openstereo}. The models are trained on various comparative datasets and evaluated on the DrivingStereo~\cite{yang2019drivingstereo} dataset.

\noindent\textbf{Training Details.} All models are trained on the SceneFlow dataset and our \textbf{RST-dataset}. The batch size is uniformly set to 8, and each model undergoes the same number of training steps across the various datasets. Models are then evaluated on adverse weather conditions from the DrivingStereo~\cite{yang2019drivingstereo} and SeeingThroughFog~\cite{bijelic2020seeing} dataset to assess their generalization performance.

\noindent\textbf{Evaluation Metrics.} We evaluate performance using End-Point Error (EPE), which measures average pixel-level disparity error, and the D1 metric, which quantifies the percentage of disparity outliers exceeding 3 pixels or 5\% of the true value, critical for assessing models' performance.

\subsection{Quantitative Results}

\noindent\textbf{Comparison with Stereo Matching Methods.} Comparison results are presented in Table~\ref{tab-SM-DS} and Table~\ref{tab-SM-STF}, where we compare state-of-the-art stereo matching methods on adverse weather datasets. On the DrivingStereo~\cite{yang2019drivingstereo} dataset, our method achieved the best results across all weather conditions, particularly in challenging scenarios such as rainy days, demonstrating its effectiveness in addressing stereo matching issues under adverse weather.

Additionally, in Table~\ref{tab-SM-STF}, we evaluate on the extreme adverse weather dataset SeeingThroughFog~\cite{bijelic2020seeing}, comparing powerfull zero-shot models StereoAnything~\cite{guo2024stereo} and LightStereo~\cite{guo2024lightstereo}. To validate the effectiveness of our dataset, we fine-tuned StereoBase~\cite{guo2023openstereo} using various datasets. The results demonstrate that our method largely improves stereo matching performance under adverse weather. Notably, even training existing models using only the RST-dataset, without additional model enhancements, yields substantial performance gains in adverse weather conditions.

\noindent\textbf{Comparison with Other Stereo Datasets.} As shown in Table~\ref{tab-DATA-DS}, we validated performance on generated data, showing that the model trained on RST-dataset achieves results comparable to state-of-the-art datasets like SceneFlow~\cite{mayer2016large}, KITTI~\cite{menze2015object}, and vKITTI~\cite{cabon2020vkitti2}. While performance across datasets is similar in normal scenarios, the RST-dataset demonstrates superior adaptability in adverse weather conditions, such as rain and fog, highlighting its effectiveness in enhancing generalization for real-world applications.

\begin{figure}[]
\centering
\small
\setlength\tabcolsep{1pt}
\begin{tabular}{cccc}
\includegraphics[width=0.245\linewidth]{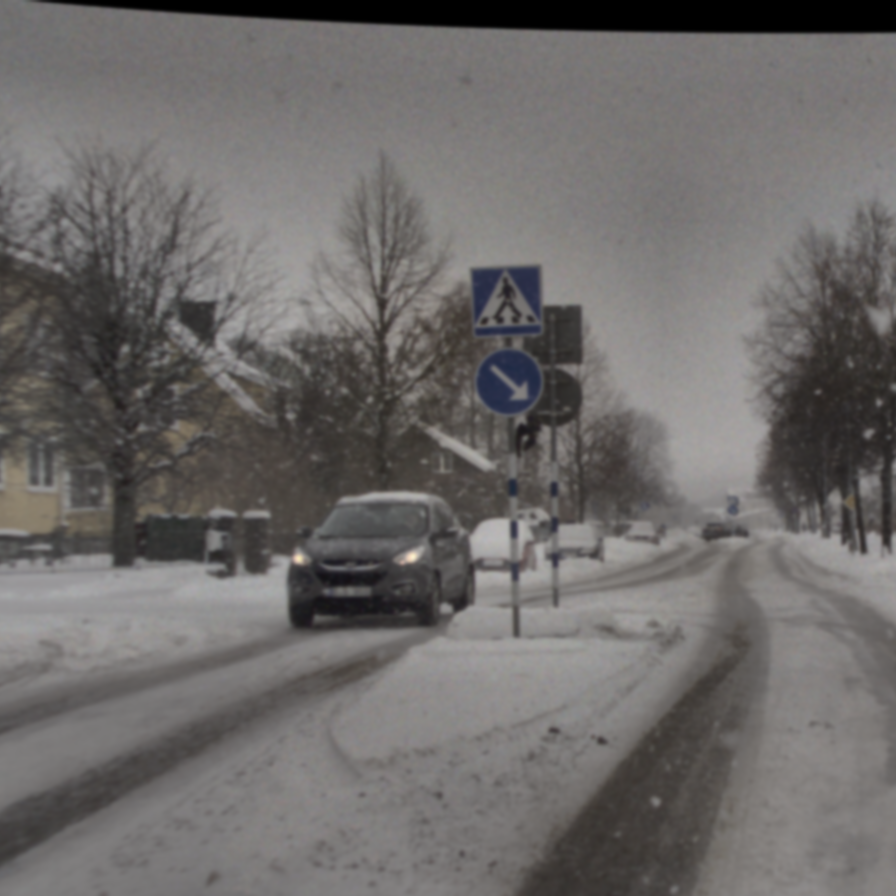} &  \includegraphics[width=0.245\linewidth]{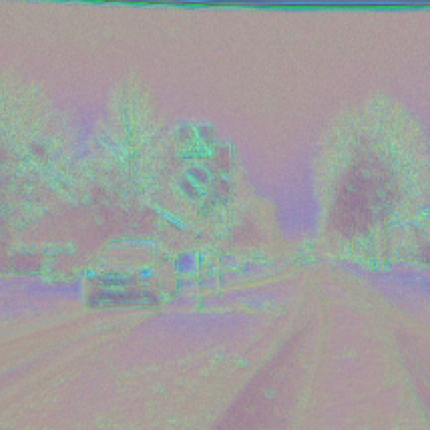} & \includegraphics[width=0.245\linewidth]{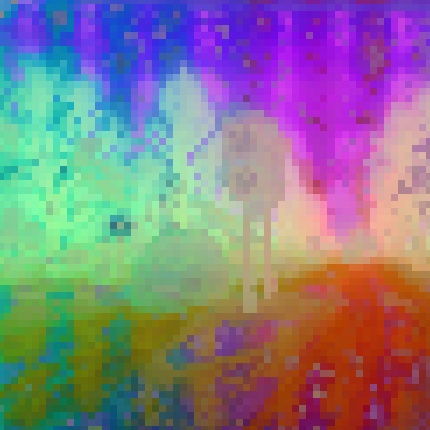} & \includegraphics[width=0.245\linewidth]{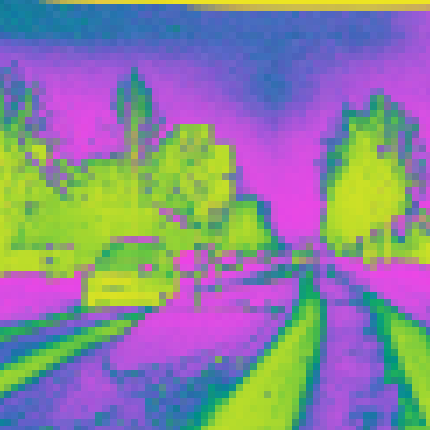} \\  \includegraphics[width=0.245\linewidth]{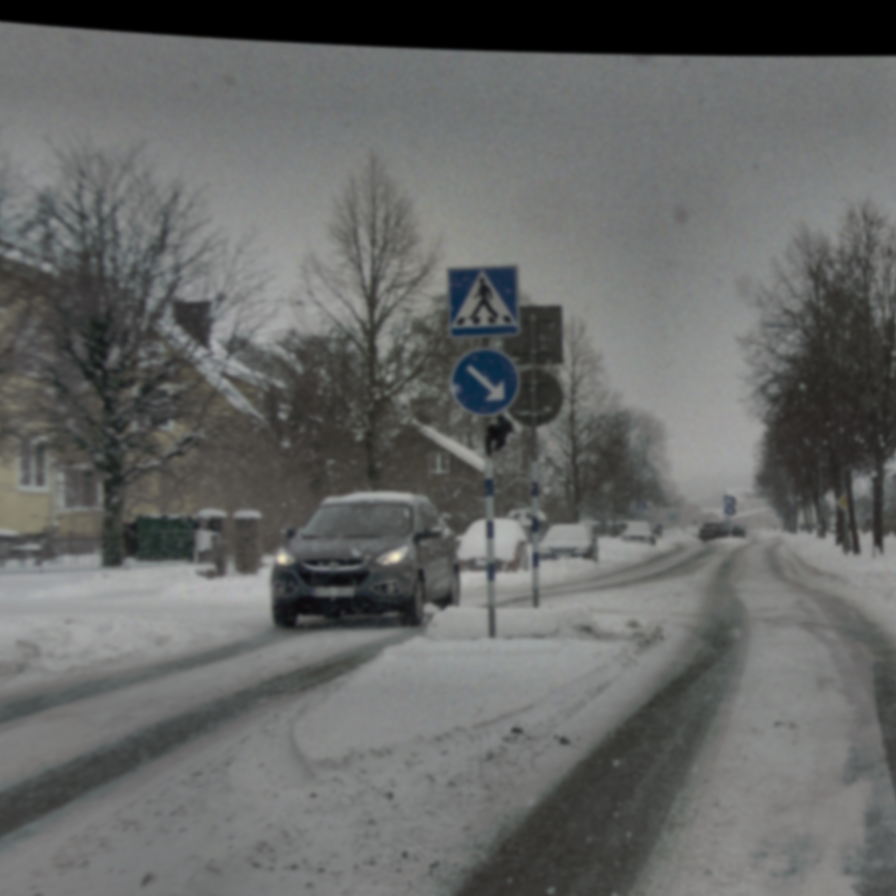} &  \includegraphics[width=0.245\linewidth]{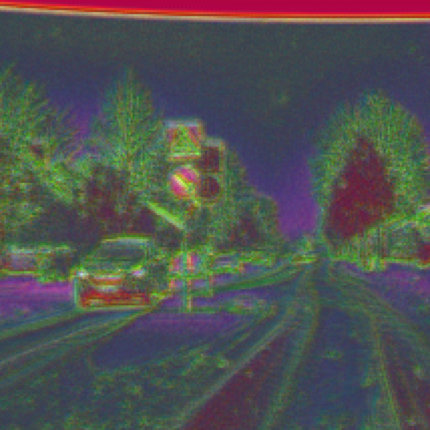} & \includegraphics[width=0.245\linewidth]{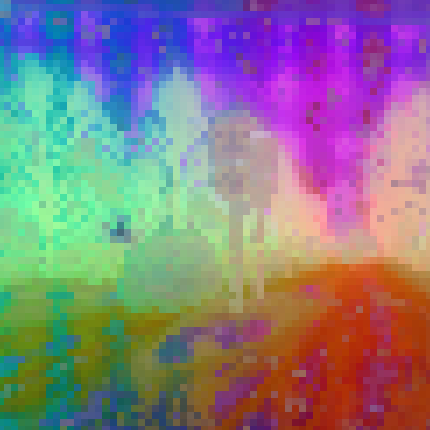} & \includegraphics[width=0.245\linewidth]{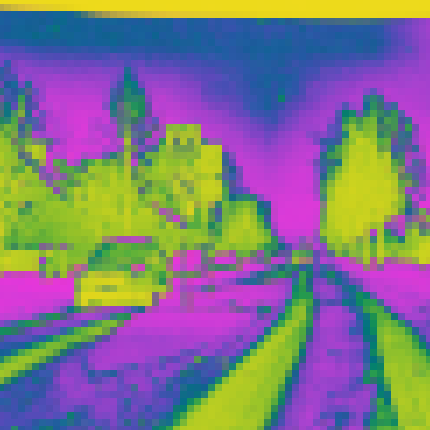} \\ 
\makecell[c]{Left/Right Img} & StereoBase~\cite{guo2023openstereo} & DinoV2~\cite{oquab2023dinov2} & Ours
\end{tabular}
\caption{Comparison of Feature Maps from Different Encoders. Feature maps are visualized using PCA~\cite{shlens2014tutorial}. Compared to other methods, our feature maps exhibit better consistency and reduced noise, improving the accuracy of subsequent stereo matching.}
\label{fig-feature}
\vspace{-1em}
\end{figure}

% \begin{figure*}
%     \centering
% \subfloat[RGB Image]{
% 		\includegraphics[width=0.09\linewidth]{img/pt-1-1.png}}
% \subfloat[SceneFlow~\cite{guo2023openstereo}]{
% 		\includegraphics[width=0.09\linewidth]{img/pt-1-2.png}}
% \subfloat[KITTI]{
% 		\includegraphics[width=0.09\linewidth]{img/pt-1-3.png}}
% \subfloat[vKITTI]{
% 		\includegraphics[width=0.09\linewidth]{img/pt-1-4.png}}
% \subfloat[Ours]{
% 		\includegraphics[width=0.09\linewidth]{img/pt-1-5.png}}
%     \caption{Qualitative comparison of stereo matching models trained on different datasets. The disparity maps and corresponding 3D point cloud reconstructions are shown in this figure. Red boxes highlight regions where DMD$^{3}$C preserves finer details and better structural integrity compared to the other methods, especially on the regions with image quality degradation under adverse weather.}
%     \label{fig:Point Cloud}
% \end{figure*}

\begin{figure*}
\centering
\small
\setlength\tabcolsep{1pt}
\begin{tabular}{ccccc}
\includegraphics[width=0.195\textwidth]{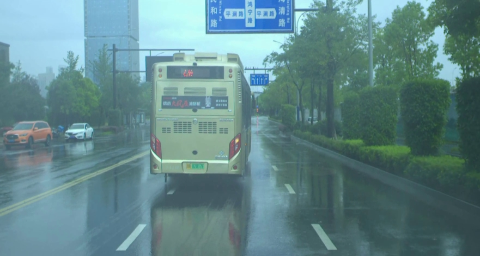} & \includegraphics[width=0.195\textwidth]{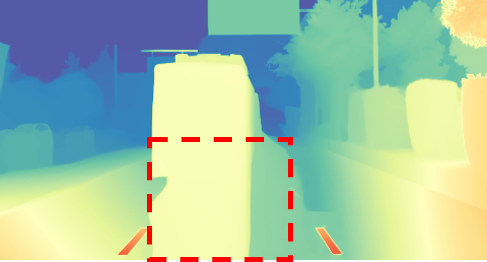} & \includegraphics[width=0.195\textwidth]{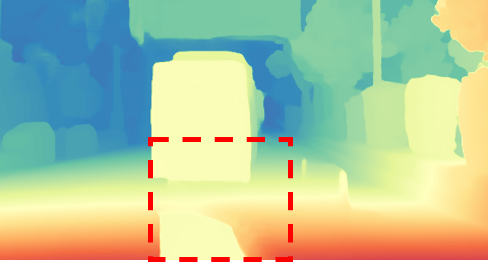} & \includegraphics[width=0.195\textwidth]{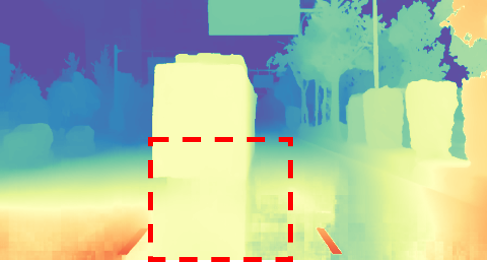} & \includegraphics[width=0.195\textwidth]{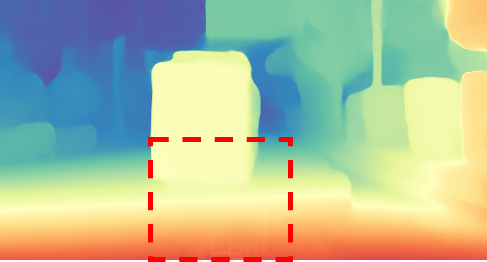} \\
\includegraphics[width=0.195\textwidth]{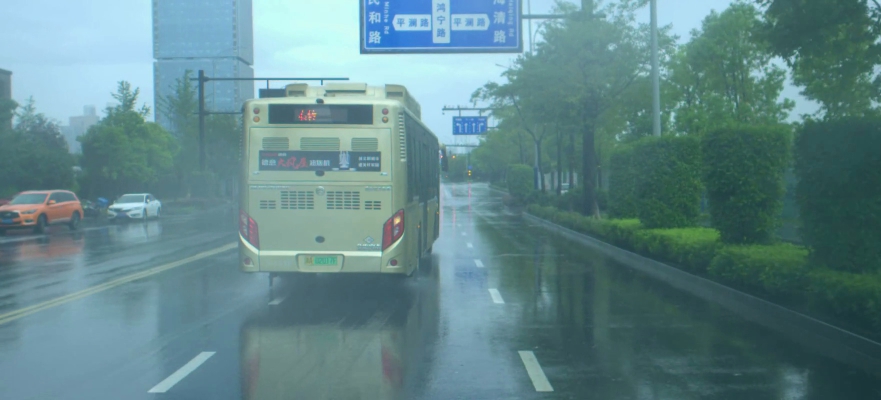} &
\includegraphics[width=0.195\textwidth]{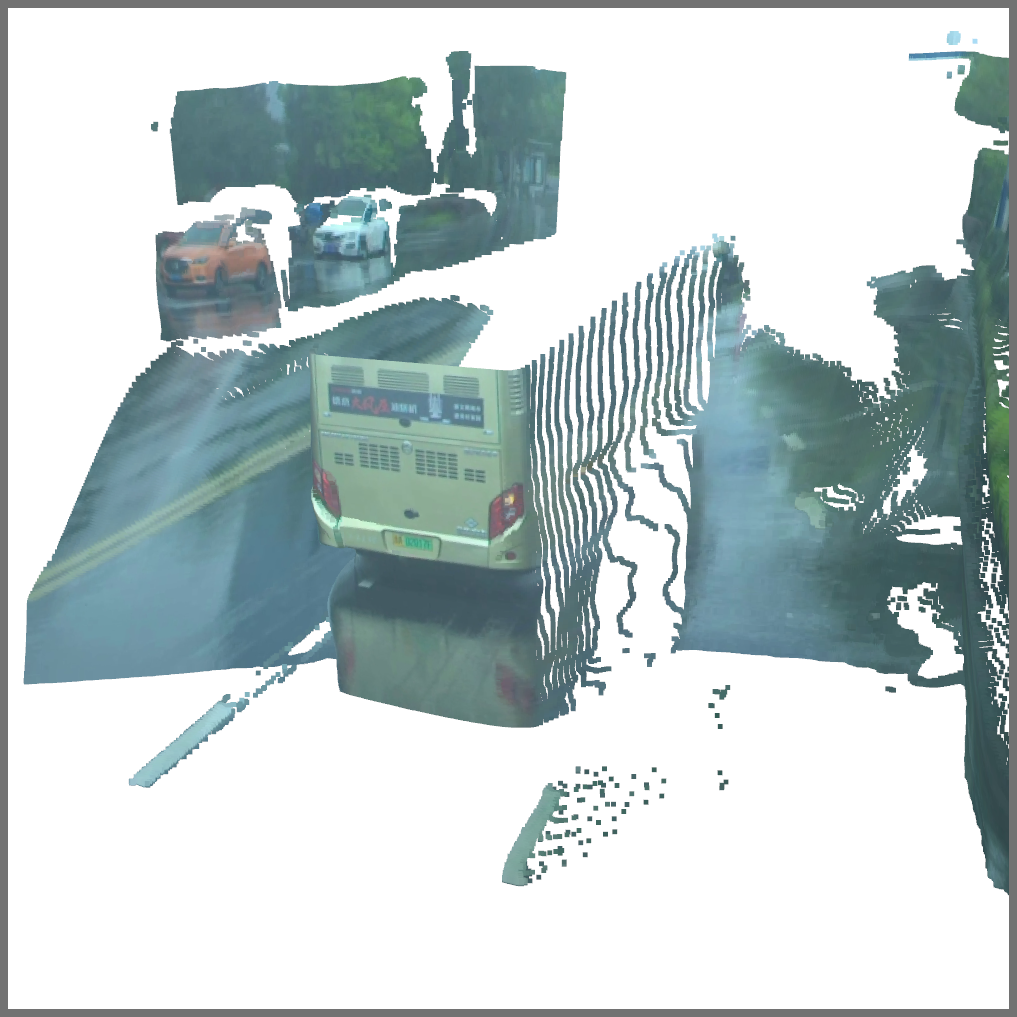} & \includegraphics[width=0.195\textwidth]{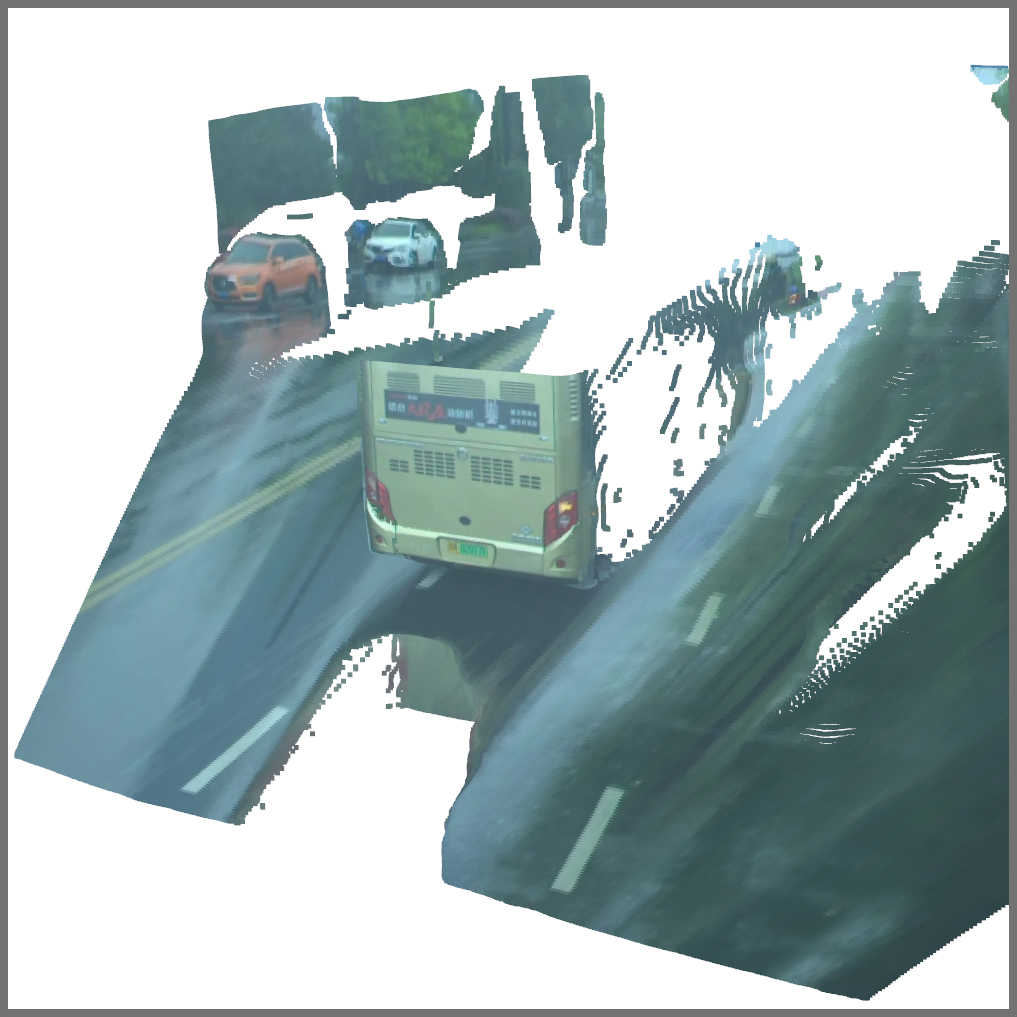} & \includegraphics[width=0.195\textwidth]{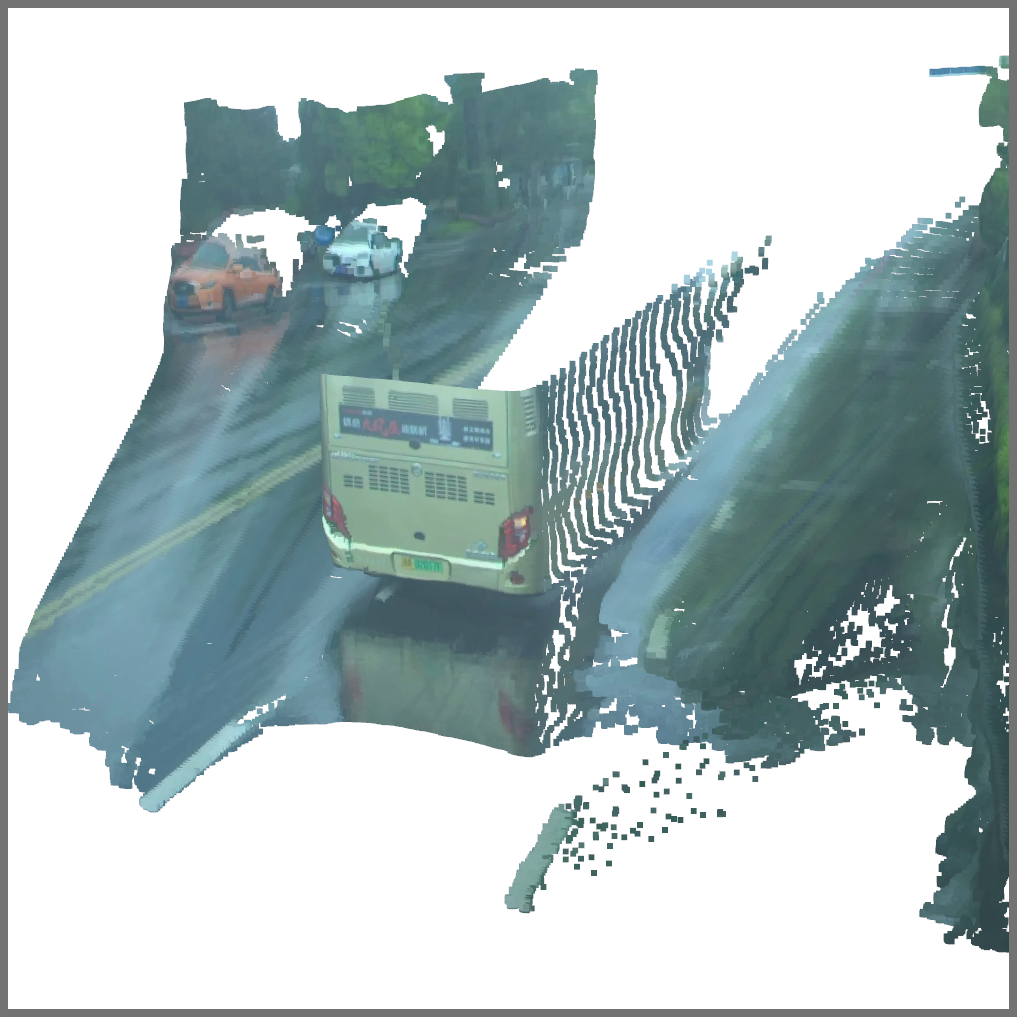} & \includegraphics[width=0.195\textwidth]{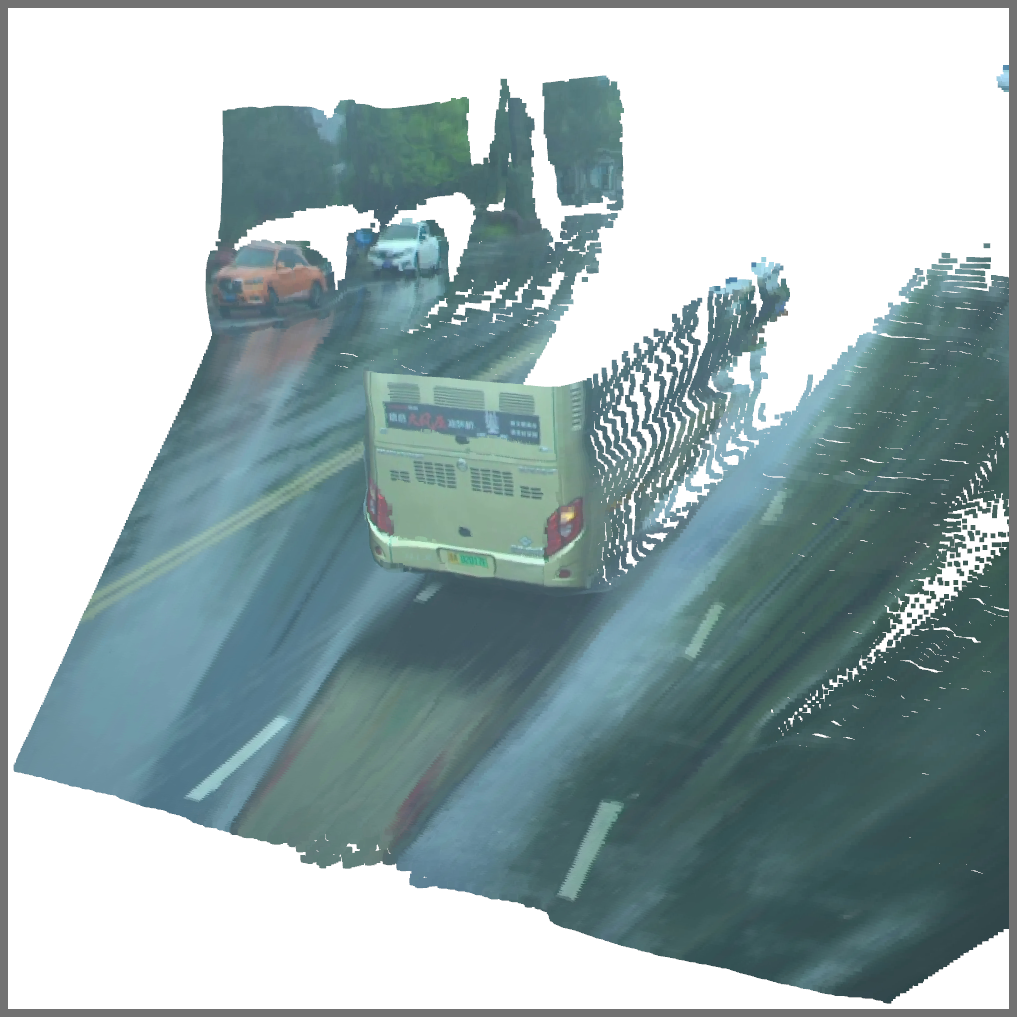} \\
Left/Right Image & StereoBase-SF~\cite{guo2023openstereo} & StereoBase-KITTI~\cite{guo2023openstereo} & StereoAnything~\cite{guo2024stereo} & Ours
\end{tabular}
\caption{Qualitative visualization of point cloud results under adverse conditions. Point clouds generated by other methods show significant artifacts on wet surfaces due to specular reflections. In contrast, our method effectively mitigates these issues, producing accurate point clouds with improved geometric consistency. A set of full-view point cloud videos are provided in the supplementary material.}
\vspace{-0.5em}
\label{fig:point-cloud}
\end{figure*}

\subsection{Qualitative Results}

\noindent\textbf{Disparity Maps in Adverse Weather.} As shown in Figure~\ref{fig:vis-dataset}, RobuSTereo outperforms other models, particularly under challenging conditions. In rainy weather, the mirror effect from wet surfaces creates challenges for stereo matching algorithms, often leading to errors in estimating ground disparity. This issue is critical for applications like autonomous driving, where inaccurate depth perception can impede obstacle detection and compromise safety. In contrast, models trained on our dataset excel on specular surfaces, providing more accurate road depth information and ensuring improved performance in such conditions.

Additionally, Figure~\ref{fig:point-cloud} compares the point cloud reconstructions from the Driving Stereo (Rainy) dataset. The improved reconstructions demonstrate the robustness of our approach in challenging weather conditions, highlighting the practical benefits of using our dataset to develop more reliable stereo vision systems for real-world applications.

\noindent\textbf{Feature Maps in Adverse Weather.} Figure~\ref{fig-feature} illustrates the feature representations extracted by different encoders under adverse weather conditions. StereoBase~\cite{guo2023openstereo} use MobileNetV2~\cite{sandler2018mobilenetv2} as encoder. DinoV2~\cite{oquab2023dinov2} denotes a standard ViT encoder. Due to the degraded image quality, noticeable noise is observed in the extracted features in MobileNetV2~\cite{sandler2018mobilenetv2} and DinoV2~\cite{oquab2023dinov2}, which impacts the subsequent matching process. Notably, our encoder effectively reduces the noise in the feature representations, enhancing the robustness of feature extraction. This improvement leads to better performance and stability of the stereo matching model in challenging weather scenarios.

\begin{table}[t]
\centering
\begin{tabular}{llcc}
\toprule
\multirow{2}{*}{Experiments} & \multirow{2}{*}{Methods} & \multicolumn{2}{c}{Driving Stereo} \\
\cmidrule(lr){3-4}
& & EPE $\downarrow$ & D1 $\downarrow$ \\
\midrule
\multirow{2}{*}{\makecell[l]{Consistency\\Module}} & Off & 1.308 & 5.997 \\
\rowcolor{graycolor} \cellcolor{white}&On & \textbf{0.875} & \textbf{2.050}  \\
\midrule
\multirow{2}{*}{Source Data} & vKITTI~\cite{cabon2020vkitti2} & 1.039 & 4.689\\
\rowcolor{graycolor} \cellcolor{white}&KITTI~\cite{menze2015object} & \textbf{0.875} & \textbf{2.050}\\
\midrule
\multirow{3}{*}{Network Encoder} & MobileNetV2~\cite{sandler2018mobilenetv2} & 0.875 & 2.050 \\
& Dinov2~\cite{oquab2023dinov2} & 0.852 & 1.793 \\
\rowcolor{graycolor} \cellcolor{white}& Robust Encoder & \textbf{0.836} & \textbf{1.598} \\
\bottomrule
\end{tabular}
\caption{Ablation experiments. Without changing other conditions such as prompt, we removed the consistency constraint module/changed the data source and generate the same number of images, using StereoBase~\cite{guo2023openstereo} as base model, and finally tested the comprehensive performance on DrivingStereo~\cite{yang2019drivingstereo}.}
\label{tab:ablations}
\vspace{-0.5em}
\end{table}

\subsection{Ablation Study}

\noindent\textbf{Coherence-Enhanced Consistency Module.} In the ablation study, we compare the proposed consistency module under same data generation config. As shown in Table~\ref{tab:ablations}, generating images without the consistency module leads to a degradation in subsequent training results. Without proper correspondence between the generated images, networks struggle to match the images correctly, which introduces confusion during the matching process.

\noindent\textbf{Data Source.} We evaluate the performance of different data sources using KITTI and vKITTI. While vKITTI offers the advantage of accurate and dense disparity ground truth (GT) and a larger dataset compared to KITTI, our experiments reveal that the performance of generated stereo images using vKITTI does not outperform that of KITTI. This discrepancy may be attributed to the relatively simple image textures in vKITTI, which differ from real-world scenes, creating a domain gap in the generated images.

\noindent\textbf{Robust Feature Encoder.} We evaluated the performance of various encoders integrated into the stereo matching model under adverse weather conditions. All experiments are conducted using the same training pipeline. The results demonstrate that our robust encoder delivers significant improvements in accuracy, which stems from encoder's ability to extract consistent and robust features, even when image quality is degraded due to adverse weather. 
% The robust encoder effectively mitigates challenges such as image noise, low visibility, and reflections, making it well-suited for challenging real-world scenarios.

\section{Conclusion}

% In this work, we proposed \textbf{RobuSTereo}, a novel framework that improves zero-shot stereo matching performance under adverse weather by tackling both data and model challenges. First, we introduced a diffusion-based data generation pipeline to synthesize high-quality stereo datasets under adverse weather scenarios, ensuring realistic and consistent stereo image pairs. Besides, we develop a coherence-enhanced consistency module to improve the consitency of the generated stereo images. Second, we enhance feature extraction stability of stereo matching model with a robust encoder, which improves the network's ability to handle degraded image quality caused by adverse weather. Both quantitative and qualitative results highlight that \textbf{RobuSTereo} substantially improves the accuracy and robustness of stereo matching models, mitigating severe errors caused by reflectivity and limited visibility. Our findings underscore the potential of data generation methods for advancing stereo data applications, particularly for scenarios that are typically underrepresented in normal datasets.

In this work, we propose RobuSTereo, a new framework to improve zero-shot stereo matching performance in adverse weather by addressing challenges in both data and models. First, we introduce a diffusion-based data generation pipeline to create high-quality stereo datasets under adverse weather, ensuring the generated stereo image pairs are realistic and consistent. We also design a coherence-enhanced consistency module to further improve the consistency of the generated images. Second, we strengthen the feature extraction of stereo matching models by using a robust encoder, which helps the model handle poor image quality caused by bad weather. Both quantitative and qualitative results show that RobuSTereo greatly improves the accuracy and reliability of stereo matching models, reducing errors caused by issues like reflections and low visibility. Our results show the potential of using advanced data generation methods to improve stereo matching, especially for scenarios that are not well-represented in existing datasets.
\section*{Acknowledgments}
This work was supported by the National Key R\&D Program of China (2022YFC3300704), the National Natural Science Foundation of China (62331006, 62171038, and 62088101), and the Fundamental Research Funds for the Central Universities.

{
    \small
    \bibliographystyle{ieeenat_fullname}
    \bibliography{main}
}

\end{document}